\documentclass[times,onecolumn,preprint,numbers,sort]{elsarticle}

\usepackage{framed,multirow}

\usepackage{amssymb}
\usepackage{latexsym}

\usepackage{url}
\usepackage{xcolor}
\definecolor{newcolor}{rgb}{.8,.349,.1}

\usepackage[ruled,linesnumbered]{algorithm2e}

\usepackage{amsmath}

\usepackage{subfigure}

\journal{Pattern Recognition Letters}

\begin{document}

\ifpreprint
  \setcounter{page}{1}
\else
  \setcounter{page}{1}
\fi

\begin{frontmatter}

\title{PSA-Det3D: Pillar Set Abstraction for 3D object Detection}

\author[sysu]{Zhicong {Huang}}

\author[sysu]{Jingwen {Zhao}}
\author[sysu]{Zhijie {Zheng}}

\author[sysu]{Dihu {Chen}\corref{cor1}}
\ead{stscdh@mail.sysu.edu.cn}

\author[sysu]{Haifeng {Hu}}

\address[sysu]{School of Electronics and Information Technology, Sun Yat-sen University, Guangzhou, China}
\cortext[cor1]{Corresponding author;}



\begin{abstract}
	Small object detection for 3D point cloud is a challenging problem because of two limitations: (1) Perceiving small objects is much more difficult than normal objects due to the lack of valid points. (2) Small objects are easily blocked which breaks the shape of their meshes in 3D point cloud. To alleviate these problems, we design a point-based detection network PSA-Det3D which mainly consists of a pillar set abstraction (PSA) and a foreground point compensation (FPC). The PSA improves the query approach of set abstraction to expand its receptive field of the network, which benefits the point-wise feature aggregation for small objects. The FPC fuses the foreground points and the estimated centers to optimize the candidate points selection. This module improves the detection performance for occluded objects effectively. The experiments on the KITTI 3D detection benchmark show that our proposed PSA-Det3D outperforms existing point‐based algorithms for small object detection.
\end{abstract}

\end{frontmatter}



\section{Introduction}
LiDAR-based 3D object detection is widely applied in autonomous driving and robotics \citep{Deep_R_Guo, A_R_Arnold, Deep_R_Cui}. Existing detectors have achieved high accuracy for car category, but the detection performance for small objects such as pedestrian and cyclist is still unsatisfactory. Thus, how to improve the detection of small objects in 3D point cloud is an important yet challenging problem \citep{Embracing_A_Fan}. Some methods apply the multimodal fusion-based approaches in the 3D object detection network \citep{Frustum_A_Wang, Range_A_Liang, Painting_A_Vora}. Due to the detailed information provided by the cameras and LiDAR sensors, they achieve high accuracy for both normal and small objects. However, the complex fusion networks increase the computational cost of these methods and thus limit their application.

Although the researches on point-based methods \citep{PointRCNN_A_Shi, 3DSSD_A_Yang, SASA_A_Chen, Not_A_Zhang} have achieved remarkable progress, the limitations of small objects are not sufficiently considered yet. There are two obvious limitations for LiDAR-based small object detection: (1) Perceiving small objects is much more difficult than normal objects because the sparse LiDAR-based point clouds usually do not provide sufficient information of small objects. However, the existing point-based methods \citep{PointRCNN_A_Shi, SASA_A_Chen, Not_A_Zhang} usually apply the native set abstraction (SA) \citep{PointNetplus_A_Qi} to encode the features, either for normal objects or small objects. (2) Small objects are easily blocked in 3D point cloud. Since the scanning beam of LiDAR sensor is not penetrating, the occlusion not only reduces the number of the valid points for small objects but also breaks the shape of their meshes. As the semantic information of the occluded objects are affected, the normal approaches \citep{PointRCNN_A_Shi, SASA_A_Chen} based on semantic features can hardly detect such objects normally. 



Considering the above limitations, we first propose a pillar set abstraction (PSA) to learn informative point-wise features for small objects. Motivated by the pillar-based methods \citep{PointPillars_A_Lang, PB_A_Wang, PiFeNet_A_Le}, our PSA adopts a pillar query operation before feature aggregation to expand the receptive field of the encoder network. Compared with the traditional SA, PSA can help to improve the performance of detection network, especially for small objects, with only a small increase on computational cost.

Furthermore, we present a foreground point compensation (FPC) to locate the objects under occlusion. In our FPC, we not only segment the foreground points but also estimate the centers of all possible objects. The centroid information can effectively compensate for incorrect segmentation results caused by occlusion. We fuse the centers and the foreground points to generate the detection proposals. Both the PSA and FPC are utilized in our point-based object detection network, PSA-Det3D. 




In summary, our main contributions include:
\begin{enumerate}[(1)]
\item We propose a pillar set abstraction to learn informative features for sparse point clouds of small objects, which adopts a pillar query operation before feature aggregation to expand the receptive field of the network.

\item We propose a foreground point compensation to locate the occluded objects. This module optimizes the selection of candidate points by the fusion of foreground points and estimated center points.

\item We propose a novel 3D object detection network, PSA-Det3D, which is evaluated on KITTI 3D object detection benchmark. Extensive experimental results confirm the outstanding accuracy of PSA-Det3D for small object detection.
\end{enumerate}

\section{Related work}
\subsection{Voxel-based Methods}
Voxel-based methods convert the irregular point clouds into dense voxels or pillars before sending them to the network. VoxelNet \citep{VoxelNet_A_Zhou} is the pioneer architecture that uses a 3D convolution network to detect 3D object from point clouds. SECOND \citep{SECOND_A_Yan} implements a fast backbone with 3D submanifold convolution network which significantly improved the efficiency of voxel-based network. In PointPillars \citep{PointPillars_A_Lang}, a pillar-based aggregation operation is proposed to obtain the pseudo-image feature map from bird's eye view (BEV). It simplifies both the voxelization process and the architecture of detection network to improve the computational efficiency. TANet \citep{TANet_A_Liu} uses triple attention mechanism in the feature encoding network to improve the detection performance for small objects and the robustness. HotSpotNet \citep{Object_A_Qi} converts the voxel-wise features into a new representation and predicts boxes with an anchor-free pipeline. Voxelization reduces the complexity and irregularity of the input data, but the generated voxels lose detailed information of the raw point clouds simultaneously, which is detrimental for small object detection. 

\subsection{Point-based Methods}
Point-based methods learn 3D features from raw point clouds directly without pre-processing. Existing point-based detectors \citep{PointNet_A_Charles, PointNetplus_A_Qi, PointRCNN_A_Shi, 3DSSD_A_Yang, 3DCenterNet_A_Qi, Pan_2021_CVPR, SASA_A_Chen, Not_A_Zhang} usually learn the point-wise features using PointNet++ and predict 3D proposals based on these features.  PointNet \citep{PointNet_A_Charles} successfully applies the neural network to aggregate the point-wise features through the multi-layer perceptron (MLP) networks. They also conduct experiments on different 3D computer vision tasks to prove the generalizability of PointNet. PointNet++\citep{PointNetplus_A_Qi} enhances the performance of feature extraction by embedding the farthest point sampling (FPS) algorithm and the grouping operation based on PointNet. PointRCNN\citep{PointRCNN_A_Shi} is a two-stage object detection algorithm proposed by Shi et al. in 2019. It adopts PointNet++ for the point-wise feature extraction and predicts 3D boxes based on the foreground point segmentation. 3D-CenterNet\citep{3DCenterNet_A_Qi} predicts the position of the object centers and estimates the bounding boxes around them. Pointformer \citep{Pan_2021_CVPR} use a local and global transformer network for feature learning in a object detection model. SASA \citep{SASA_A_Chen} present a  semantics-augmented set abstraction module to maintain more foreground points in the point-based encoder. Zhang et al. \citep{Not_A_Zhang} implement a fast single-stage detector IA-SSD by using an instance-aware downsampling strategy to select foreground points of interest and predict the centroids of objects. 

\subsection{Point-Voxel Fusion Methods}
In addition to the above methods, there is another stream that uses the voxel-based and the point-based networks in a single detector \citep{FastPR_A_Yilun, PVRCNN_A_Shi, SASSD_A_He, Spatial_A_Ziyu, BADet_A_Rui}. Fast PointRCNN \citep{FastPR_A_Yilun} adopts a point-based network to fuse the voxel-wise features from a voxel-based backbone for the optimization of 3D proposals. In PV-RCNN \citep{PVRCNN_A_Shi}, the voxel-wise features are mapped to each sampled key point through the voxel set extraction module where the obtained point-wise features are further used to optimize the proposals. SA-SSD \citep{SASSD_A_He} projects the voxel-wise features back to the point-wise features to exploit point-wise supervisions in an auxiliary network. The auxiliary network guides the voxel-wise features in the backbone network to be aware of the object structure, thus improving the detection performance while introducing no extra computation in the inference stage. M3DETR \citep{M3DETR_A_2022} learns multi-representation features of point cloud and fuse the features using a multi-scale transformer network. The point-voxel fusion methods contribute to a better detection performance, but how to effectively fuse the features of the two modalities is still an open and challenging problem.

\begin{figure*}[t]
\centering
\includegraphics[width=0.9\textwidth]{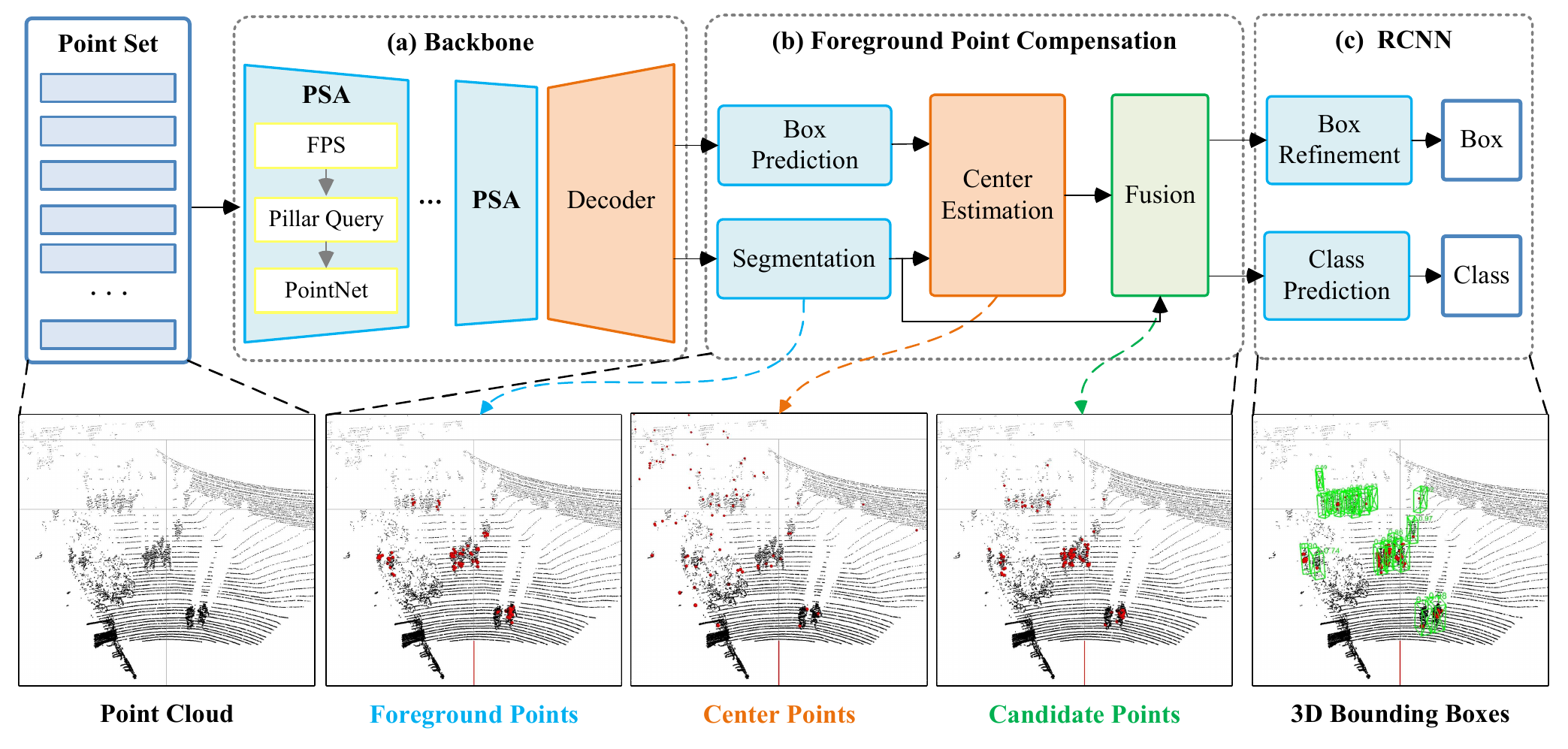}
\caption{Outline of PSA-Det3D. The input points pass through (a) backbone network with PSA layer to produce the point-wise features. The (b) foreground point compensation is used to predict 3D proposals based on the features. The final detection results are generated in the (c) RCNN.}
\label{fig:psa-det3d}
\end{figure*}

\section{Our Method}
As shown in Fig. \ref{fig:psa-det3d}, PSA-Det3D mainly comprises three components: (a) a point-based backbone with PSA layers; (b) a foreground point compensation (FPC) block; and (c) an RCNN which generates the final detection results. The input point set $P$ first passes through the backbone to produce the point-wise features $F$. In this block, several stacked PSA layers are utilized to build the feature encoder. Then the point-wise features are sent to the FPC block. We predict 3D bounding boxes $BB$ and calculate the foreground confidence by a segmentation following PointRCNN \citep{PointRCNN_A_Shi} in the FPC first. Then we employ a center estimation as well as a fusion algorithm to optimize the candidate points and generate 3D proposals. At last, the proposals are refined by an RCNN. The details of the PSA and FPC will be described below.

\subsection{Pillar Set Abstraction} 

\begin{figure}[!b]
\centering
\includegraphics[width=0.5\textwidth]{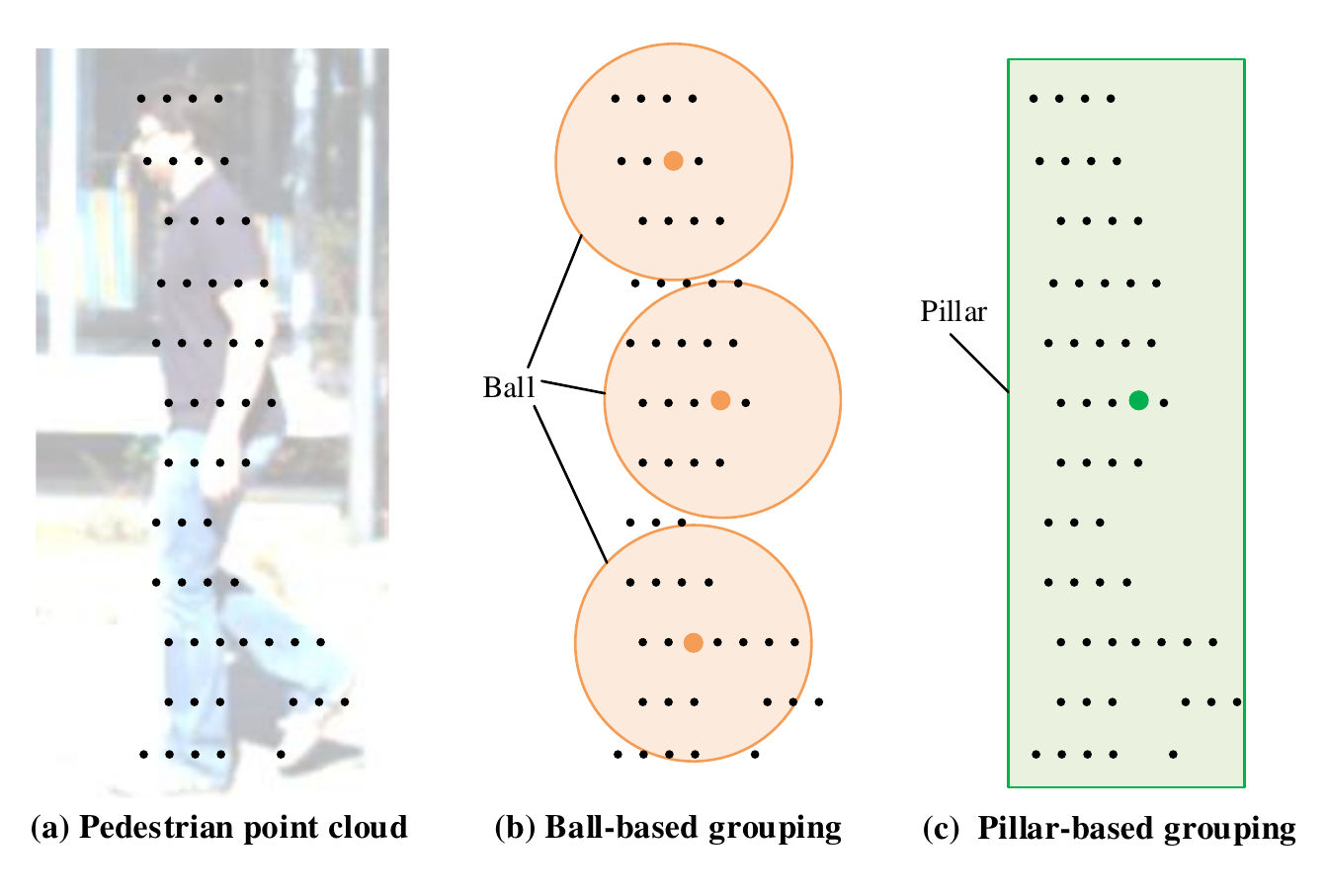}
\caption{Both (b) ball query operation and (c) pillar query operation are simulated on the (a) point cloud of a pedestrian.}
\label{fig:group_operation}
\end{figure}

The SA of PointNet++ is widely applied to learn the point-wise features in existing point-based methods \citep{PointRCNN_A_Shi, 3DSSD_A_Yang, 3DCenterNet_A_Qi, SASA_A_Chen, Not_A_Zhang}. In SA, the input points are divided into several balls around the sampled key points in 3D space through a ball query operation \citep{PointNetplus_A_Qi}. We use the point cloud of a pedestrian as an example to simulate this operation. As shown in Fig. \ref{fig:group_operation}, the ball query of SA generates balls around the key points to aggregate their local features. However, this operation is restricted by the number of sampled points and the ball radius adopted \citep{3D_A_Mao}. If the number of the key points reduces to two or even one, the ball-based grouping operation cannot keep a high coverage anymore. This is probably one of the main reasons for the decreased accuracy on small objects because of insufficient context information. Simply increasing the radius of the balls cannot solve the problem because it would introduce severe noise to the aggregated features and the fine-grained 3D information may lose. 

The pillar-based methods \citep{PointPillars_A_Lang, PB_A_Wang} provide a special query operation that partitions points into pillars by increasing the boundaries of voxels in the z-direction. Motivated by this, we propose a novel pillar set abstraction (PSA) which implements the pillar-based grouping operation in the SA of PointNet++ \citep{PointNetplus_A_Qi}. Specifically, we use the FPS algorithm to sample key points and PointNet to learn the features in the PSA layer. After the FPS, we employ a pillar-based grouping operation to partition points into several groups. This operation is represented by the following formula:


\begin{equation}
\begin{aligned}
	\label{eq:ga}
	&GA(k) = \{p \in P \mid HD(p,k) < r_0\}, k \in K \\
	&HD(p,k) = \sqrt{(p(x) - k(x))^2 + (p(y) - k(y))^2} \\
	&P = \{p_n\}^{N_i-1}_{n=0}, K=\{k_n\}^{N_o-1}_{n=0}, N_o \leq N_i
\end{aligned}
\end{equation}

\noindent where $GA$ is the array of pillar-based grouping result, $P$ is the set of input points, $N_i$ is the number of input points, $K$ is the set of key points, $N_o$ is the number of sampled points,  $HD(p,k)$ is the horizontal distance between points $p$ and $k$. 

The pillar-based grouping of our PSA is also simulated for comparison. As shown in Fig. \ref{fig:group_operation}, the generated pillar captures most of the points even if there is only one key point used. In other words, the PSA provides more valid points to the MLPs and thus helps to learn informative features.

\subsection{Foreground Point Compensation}
After feature learning, the point-based detector needs to select candidate points for 3D proposal generation. Existing methods \citep{PointRCNN_A_Shi, Pan_2021_CVPR, SASA_A_Chen} usually adopt the foreground point segmentation to select the most valuable points. However, the foreground scores cannot fully reflect the quality of the predicted proposals \cite{PointGNN_A_Shi} because the relationship between 3D proposals and points is not considered. Additionally, the occlusion of small objects will inevitably affects the accuracy of the segmentation due to the lack of semantic information. An increasing number of approaches produce the candidate points by locating the object centers \citep{Deep_A_Qi, 3DCenterNet_A_Qi, Not_A_Zhang}, which is an effective improvement compared with the foreground point segmentation. Therefore, our proposed FPC fuses both the foreground points and the object centers to optimize the selection of candidate points. As shown in Fig. \ref{fig:psa-det3d}, we adopt a box prediction and segmentation to predict preliminary 3D proposals and semantic scores. Next, we present a center estimation algorithm to predict object centers, which are then used for candidate points selection in a fusion module. We detail the center estimation in algorithm \ref{alg1}.


\begin{algorithm}[!h]
\caption{Center Estimation algorithm}
\label{alg1}
\KwIn{the size of input points: $N$; the input points: $P = \{p_n\}^{N-1}_{n=0}$; the foreground confidences: $C_{FG} = \{C_i\}^{N-1}_{n=0}$; the set of bounding boxes: $BB = \{bb_n\}^{N-1}_{n=0}$\;}
\KwData{Center point mask: $M_{center}[0...N-1]$; Ignored point mask: $M_{ignored}[0...N-1]$}
{Initialize $M_{center}$ with zeros\;}
{Initialize $M_{ignored}$ with zeros\;}
\For{$i=0$ to $N-1$}{
	{$idx = argmax(C_{FG})$\;}
	{$C_{FG}[idx] = 0$\;}
	\If{$M_{center}[idx] == 0$ and $M_{ignored}[idx] == 0$}{
		\For{$j = 0$ to $N-1$}
		{
			\If{$P[j] \in BB[idx]$}
			{$M_{ignored}[j] = 1$\;}
		}
		$M_{center}[idx] = 1$\;
	}
}
\Return{$M_{center}$}
\end{algorithm}

The inputs include the size of input points $N$, the foreground confidences $C_{FG}$ and the set of generated 3D bounding boxes $BB$. The mask of centers $M_{center}$ is calculated based on the intrinsic relationship between the predicted 3D boxes and the coordinates of the point cloud. First, the point with the highest foreground confidence is selected as the first center $t_0$. The corresponding 3D box $BB[t_0]$ is used to identify the other points in the same box which are marked as ignored points in the mask $M_{ignored}$. Then, we continue to select the next unmarked point with the highest confidence, and perform the same operation until all points have been marked as centers or ignored points.

To generate the refined scores of 3D proposals, we design an algorithm to fuse the center mask $M_{center}$ with the foreground confidences $C_{FG}$. The final score generated through the FPC $C_{FPC}$ is formulated as follows:
\begin{equation}
\label{eq:fusion}
C_{FPC} = C_{FG} + r \cdot C_{FG} \cdot M_{center}
\end{equation}

\noindent where $r$ is a hyperparameter that determines the fusion ratio. 

\begin{figure}[!t]
\centering
\includegraphics[width=0.5\textwidth]{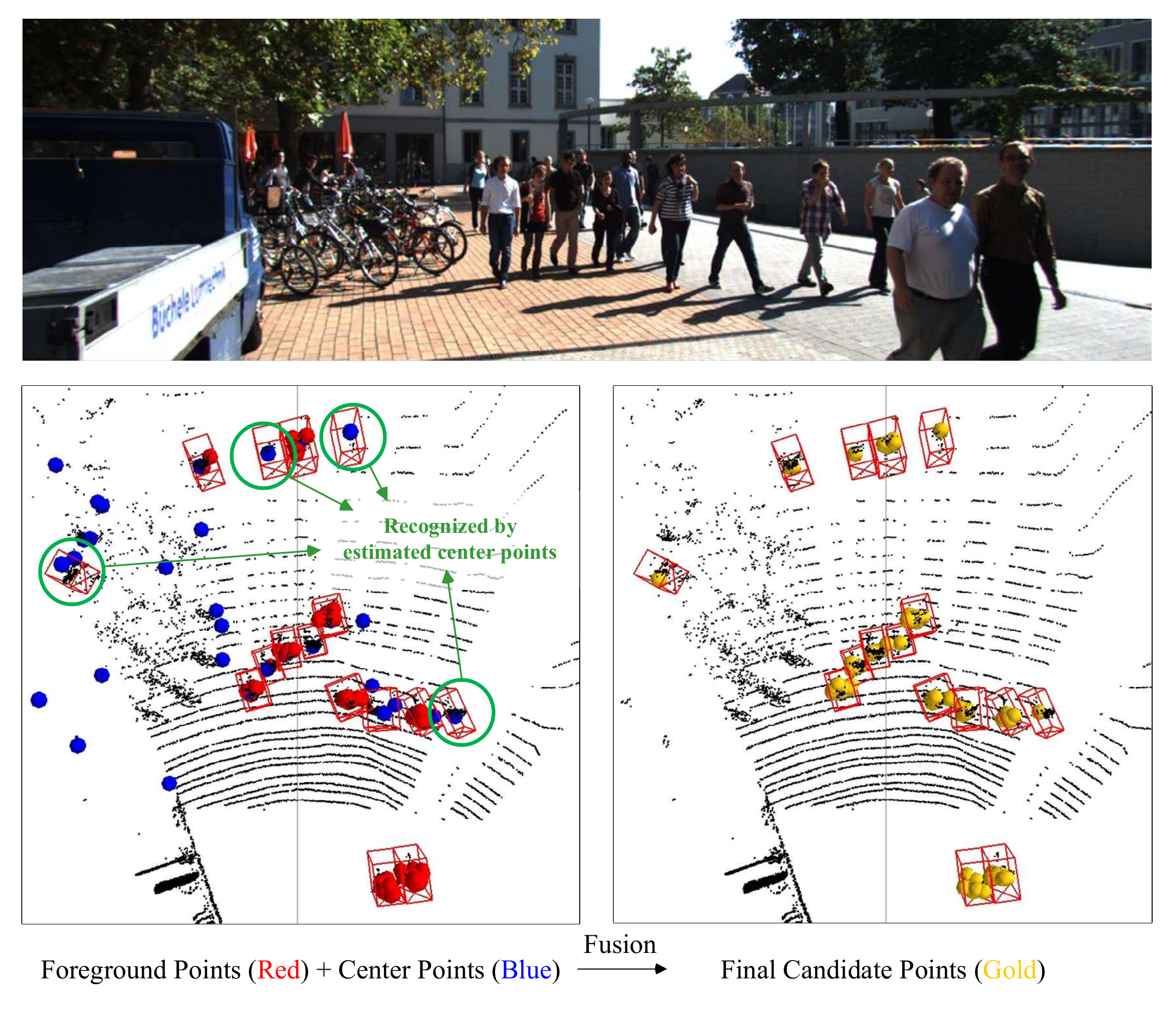}
\caption{Illustration of the FPC. We fuse the foreground points and the estimated center points to generate the optimized candidate points.}
\label{fig:FPC}
\end{figure}

As shown in Fig. \ref{fig:FPC}, we demonstrate the selection of candidate points during the FPC, where the groundtruth boxes are represented by red boxes. As the red points represent the selection of the segmented foreground points, we can see that some occluded pedestrians are not detected by the segmentation due to the occlusion in 3D point cloud. These pedestrians are then recognized by the estimated centers which are drawn as blue points. Finally, our FPC generates the proper candidate points by fusing the foreground points as well as the estimated center points, which are illustrated as gold points.

\begin{table*}[!t]
\centering
\caption{Performance comparison of 3D object detection with existing methods on KITTI $test$ split. All results are evaluated by mean Average Precision with 40 recall positions via the official KITTI evaluation server.}
\label{tab:comparison_on_kitti_test}
\resizebox{0.8\textwidth}{!}{
	\begin{tabular}{c|ccc|ccc|ccc}
		\hline
		\multirow{2}{*}{Methods} &  \multicolumn{3}{c}{Car 3D AP (\%)} \vline &  \multicolumn{3}{c}{Ped. 3D AP (\%)} \vline& \multicolumn{3}{c}{Cyc. 3D AP (\%)}  \\
		& Easy & Moderate & Hard & Easy & Moderate & Hard & Easy & Moderate & Hard \\
		\hline
		\hline
		VoxelNet \citep{VoxelNet_A_Zhou} & 77.47 & 65.11 & 57.73 & 39.48 & 33.69 & 31.50 & 61.22 & 48.36 & 44.37  \\
		SECOND \citep{SECOND_A_Yan} & 84.65 & 75.96 & 68.71 & 45.31 & 35.52 & 33.14 & 75.83 & 60.82 & 53.67  \\
		PointPillars \citep{PointPillars_A_Lang} & 82.58 & 74.31 & 68.99 & \bf{51.45} & 41.92 & 38.89 & 77.10 & 58.65 & 51.92  \\
		PointRCNN \citep{PointRCNN_A_Shi} & 86.96 & 75.64 & 70.70 & 47.98 & 39.37 & 36.01 & 74.96 & 58.82 & 52.53  \\
		Pointformer \citep{Pan_2021_CVPR} & 87.13 & 77.06 & 69.25 & 50.67 & 42.43 & \bf{39.60} & {75.01} & {59.80} & {53.99}  \\
		IA-SSD \citep{Not_A_Zhang} & 88.34 & 80.13 & 75.04 & 46.51 & 39.03 & 35.61 & \bf{78.35} & \bf{61.94} & \bf{55.70}  \\
		SASA \citep{SASA_A_Chen} & \bf{88.76} & \bf{82.16} & \bf{77.16} & - & - & - & - & - & - \\
		\hline
		{PSA-Det3D} & 87.46 & 78.80 & 74.47 & 49.72 & \bf{42.81} & {39.58} & 75.82 & 61.79 & 55.12  \\
		\hline
	\end{tabular}
}
\end{table*}

\begin{table*}[!t]
\centering
\caption{Quantitative comparison of different approaches on the $val$ split of the KITTI dataset. Our PSA-Det3D achieves the best performance for small objects including ``pedestrian" and ``cyclist".}
\label{tab:comparison_on_kitti_val}
\resizebox{0.8\textwidth}{!}{
	\begin{tabular}{c|ccc|ccc|ccc}
		\hline
		\multirow{2}{*}{Methods} & \multicolumn{3}{c}{Car 3D AP (\%)} \vline & \multicolumn{3}{c}{Ped. 3D AP (\%)} \vline& \multicolumn{3}{c}{Cyc. 3D AP (\%)} \\
		& Easy & Moderate & Hard & Easy & Moderate & Hard & Easy & Moderate & Hard \\
		\hline
		\hline
		VoxelNet \citep{VoxelNet_A_Zhou} & 81.97 & 65.46 & 62.85 & 57.86 & 53.42 & 48.87 & 67.17 & 47.65 & 45.11  \\
		SECOND \citep{SECOND_A_Yan} & 90.55 & 81.61 & 78.61 & 55.94 & 51.14 & 46.17 & 82.96 & 66.74 & 62.78  \\
		PointPillars \citep{PointPillars_A_Lang} & 87.75 & 78.40 & 75.18 & 57.30 & 51.41 & 46.87 & 81.58 & 62.94 & 58.98 \\
		PointRCNN \citep{PointRCNN_A_Shi} & 92.07 & 80.61 & 78.35 & 64.01 & 57.34 & 51.10 & 92.10 & 73.03 & 68.52  \\
		Pointformer \citep{Pan_2021_CVPR} &90.05 & 79.65 & 78.89 & - & - & - & - & - & -\\
		IA-SSD \citep{Not_A_Zhang} &91.88 & 83.41 & 80.44 & 61.22 & 56.77 & 51.15 & 88.42 & 70.14 & 65.99\\
		SASA \citep{SASA_A_Chen} & \bf{92.19} & \bf{85.76} & \bf{83.11} & 67.46 & 61.52 & 55.75 & 91.57 & 73.89 & 69.63 \\
		\hline
		PSA-Det3D & 91.34 & 80.53 & 78.09 & \bf{68.35} & \bf{63.09} & \bf{57.09} & \bf{92.23} & \bf{74.11} & \bf{69.97}  \\
		\hline
	\end{tabular}
}
\end{table*}

\subsection{Refinement and loss function}
We adopt the box refinement layer and the loss function from \citep{PointRCNN_A_Shi}. The 3D bounding box regression loss $L_{reg}$ can be formulated as
\begin{equation}
\begin{aligned}
	\label{eq:loss-1}
	& L_{reg} = \frac{1}{N_{pos}} \sum_{p \in pos}(L_{bin}^{(p)} + L_{res}^{(p)}) \\
\end{aligned}
\end{equation}

\noindent where $N_{pos}$ is the number of foreground points, $L_{bin}^{(p)}$ is the bin-based regression losses and $L_{res}^{(p)}$ is smooth L1 loss of regression.

\section{Experiments}
Our 3D detector is evaluated on the KITTI object detection benchmark \citep{KITTI_A_Geiger} which provides 7481 training samples and 7518 test samples. The objects are labeled as three levels of difficulty (``easy", ``moderate" and ``hard"), according to their occlusion level and truncation ratio. Based on the common practice in related research, we divide training data into ``train" split (3712 samples) and ``val" split (3769 samples) for training and evaluating detection networks.


\subsection{Implementation Detail}
Our PSA-Det3D is implemented based on the OpenPCDet framework \citep{openpcdet2020}. As only objects appearing on the image plane are labeled, the point clouds of KITTI dataset are first filtered through a projection from 3D to 2D. Next, we randomly sample 16384 points from the remaining points as the input of our detector. In terms of network parameters, the sizes of key points in four PSA layers are set to 8192, 1024, 256 and 64 respectively. The receptive radius of the PSA layers is set to 0.1, 0.5, 1.0 and 2.0 in meters. We implement the FPC module in the proposal generation layer after the foreground point segmentation and the 3D box prediction. The fusion ratio $r$ of formula \ref{eq:fusion} is set to 0.5. 

During training, we adopt three commonly used data augmentation methods, including random flipping about the X axis or Y axis, random scaling, and random rotation about the Z axis. The network is trained on 2 TITAN Xp GPUs for total 80 epochs with Adam optimizer \citep{Super_A_Leslie} and one cycle learning rate schedule.

\subsection{Online evaluation on test set}
We evaluate the detection performance of our method and several other methods from existing literatures \citep{VoxelNet_A_Zhou, SECOND_A_Yan, PointPillars_A_Lang, PointRCNN_A_Shi, SASA_A_Chen, Not_A_Zhang} on KITTI $test$ split dataset for ``car", ``pedestrian" and ``cyclist" under three different difficulties. The results on ``moderate" are usually adopted as the main indicator for final ranking. Note that, since \citep{SASA_A_Chen} does not submit the result for ``pedestrian" and ``cyclist" to KITTI server, we only report its performance for ``car".  For the most challenging ``pedestrian" detection race track, our method achieves 42.81\% ``moderate" 3D AP, outperforming Pointformer \citep{Pan_2021_CVPR} by 0.38\%. Compared with the most recent point-based detector IA-SSD \citep{Not_A_Zhang}, we improve the 3D AP of ``pedestrian" category by 3.21\%, 3.78\% and 3.97\% on three difficulties.

\begin{figure*}[!t]
\centering
\includegraphics[width=1\textwidth]{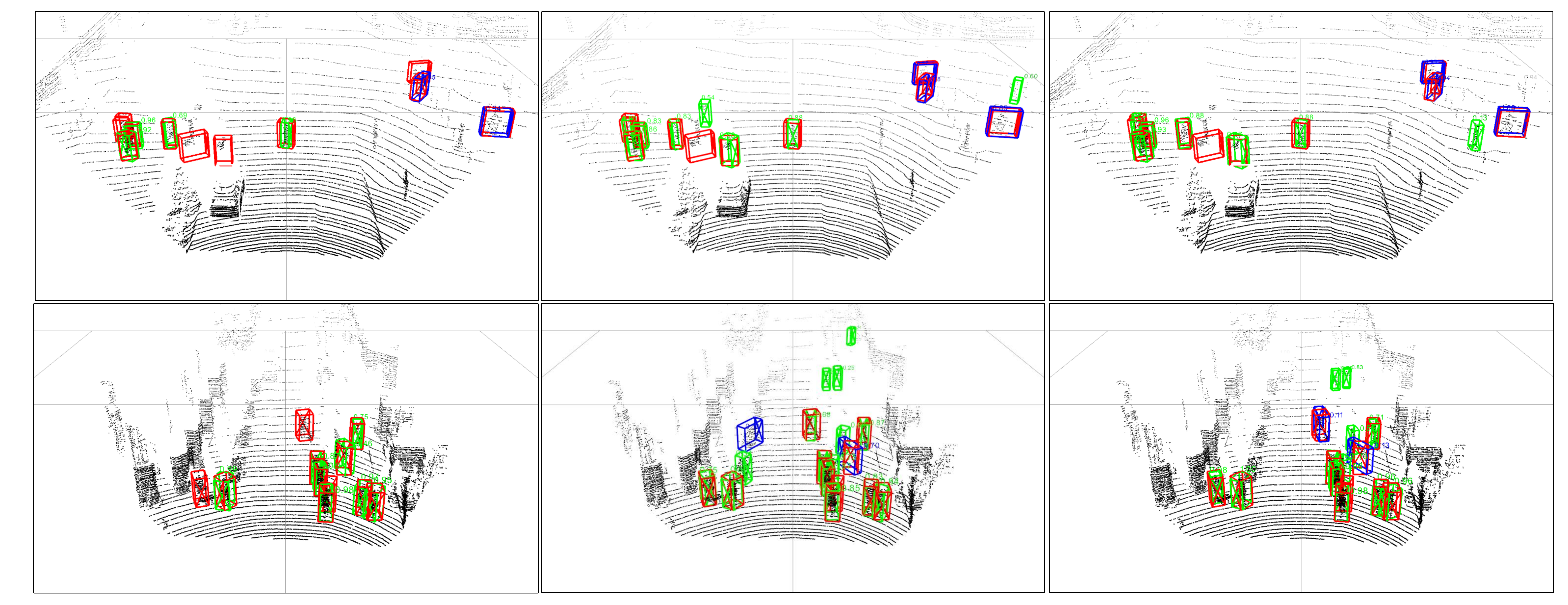}
\caption{The detection result PointRCNN ($left$), SASA ($middle$) and PSA-Det3D ($right$). The green and blue boxes indicate the predicted pedestrians and cyclists respectively while the groundtruth boxes are represented by the red boxes.}
\label{fig:visualization}
\end{figure*}

\subsection{Offline evaluation on val split}
Apart from the detection results on the $test$ split, we also report the performance comparison on the $val$ split of the KITTI dataset in Table \ref{tab:comparison_on_kitti_val}. It can be seen that our method achieves the best performance for both ``pedestrian" and ``cyclist". Comparing with the baseline model PointRCNN, our PSA-Det3D boosts the AP by 3.34\%, 5.75\%, 5.99\% for ``pedestrian" and 0.13\%, 1.08\%, 1.45\% for ``cyclist" respectively. Since we mainly focus on small object detection in this paper, the performance for ``car" is basically the same as that of our baseline model.


We visualize some representative results using different methods in Fig. \ref{fig:visualization}. We can observe that PointRCNN \citep{PointRCNN_A_Shi} locates most of the objects in the point clouds but there are still some unrecognized objects, which means that the accuracy of PointRCNN on small object detection is unsatisfactory. SASA \citep{SASA_A_Chen} generates plenty of 3D proposals to figure out more objects, but some incorrect boxes are introduced simultaneously. The performance of PSA-Det3D ($right$) is generally the best since it identifies more objects than PointRCNN while producing fewer false boxes than SASA.

\subsection{Ablation Study}
To analyze the effectiveness of the proposed PSA and FPC, we conduct comprehensive ablation studies. All models are trained on the $train$ set and evaluated on the $val$ set for ``pedestrian" and ``cyclist" categories of KITTI dataset \citep{KITTI_A_Geiger}.

\begin{figure}[!b]
\centering
\includegraphics[width=0.5\textwidth]{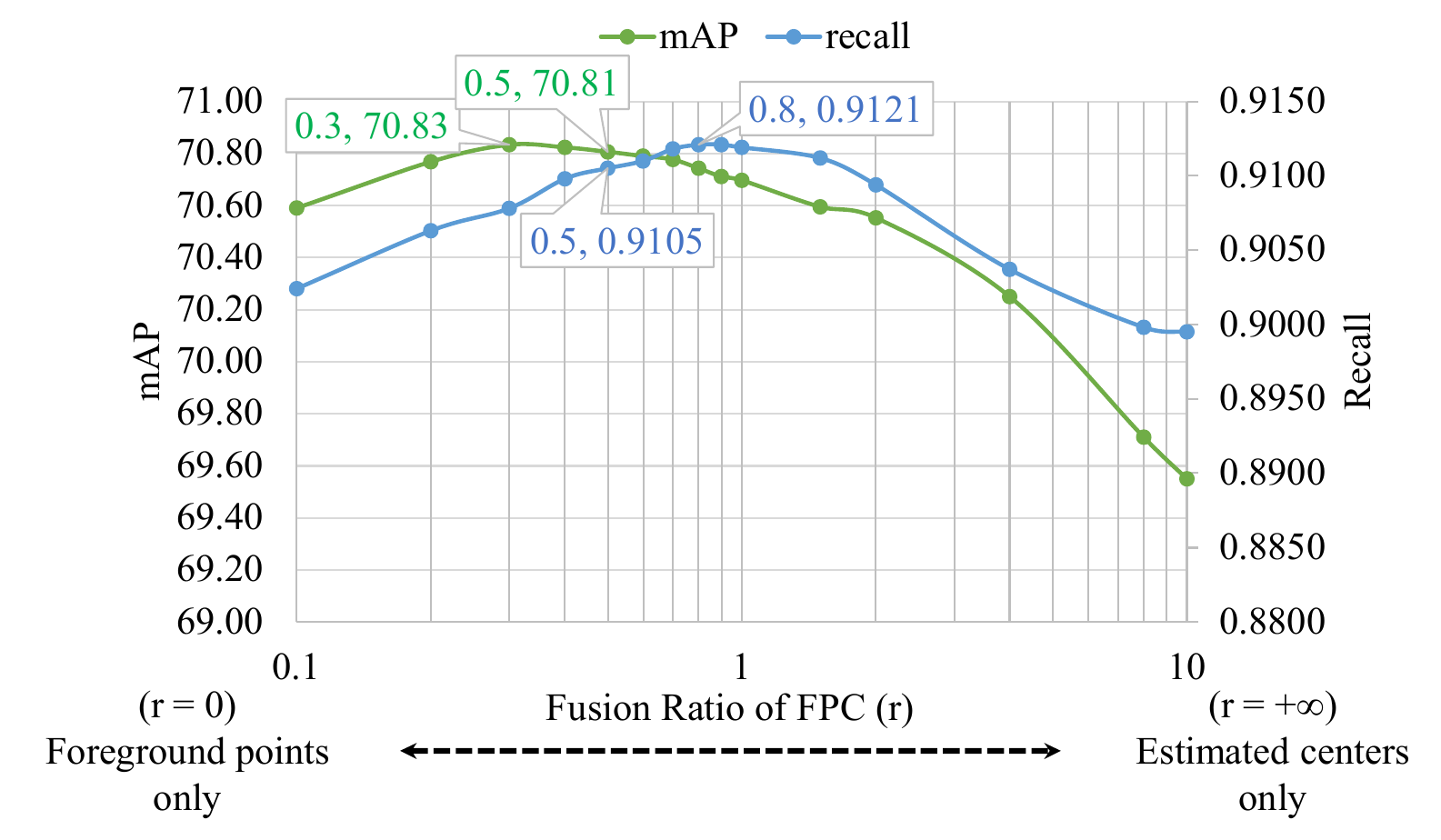}
\caption{Comparison of different fusion ratio $r$ in FPC.}
\label{fig:fusion_ratio}
\end{figure}

\begin{table}[!h]
\centering
\caption{Effectiveness of the PSA layer for small objects with different numbers of key points.}
\label{tab:sa_vs_psa}
\resizebox{0.5\textwidth}{!}{
	\begin{tabular}{c|c|c||c|c|c}
		\hline
		\multirow{2}{*}{Methods} & \multirow{2}{*}{Query} & {Number of} & {3D mAP for} & {Batch} & \multirow{2}{*}{FPS} \\
		{} & & {key points} & {$Ped.$ and $Cyc.$(\%)} & {sizes} & {} \\
		\hline
		\hline
		\multirow{2}{*}{PointRCNN \citep{PointRCNN_A_Shi}} & \multirow{2}{*}{Ball} &  4096 & 67.70 & 21 & 8.97 \\
		& & 8192 & 67.72 & 20 & 7.85 \\
		\hline
		\multirow{2}{*}{+PSA} & \multirow{2}{*}{Pillar} &  4096 & 69.11 & 20 & 8.57 \\
		& & 8192 & 70.43 & 19 & 7.69 \\
		\hline
	\end{tabular}
}
\end{table}

\subsubsection{{Effect of Pillar Set Abstraction}}

To investigate the effectiveness of the proposed PSA, we replace the encoder network of the PointRCNN with PSA and evaluate their performance and computational cost on KITTI dataset. We report the maximum number of batches that can be parallelized on one TITAN Xp(12GB) as well as the inference speed. As shown in Table \ref{tab:sa_vs_psa}, the methods ``+PSA" achieves higher accuracy than the baseline model with only a small increase on computational cost. It is worth noting that when the number of key points in the first abstraction layer increases to 8192, the performance of PSA-based method is further boosted by 1.32\%. 

The results in Table \ref{tab:ablation_ped_cyc} show that the PSA improves the detection performance for small object by 2.73\% on 3D mAP. This is mainly because the native point-wise encoder of PointRCNN is difficult to perceive small objects, especially the pedestrian. Note that our PSA significantly improves the 3D AP for ``pedestrian" by 4.71\%, 4.79\% and 4.17\% on three difficulties, which indicates that our PSA aggregates more informative features for small objects.

\begin{table}[!h]
\centering
\caption{Ablation study on the PSA and the FPC.}
\label{tab:ablation_ped_cyc}
\resizebox{0.5\textwidth}{!}{
	\begin{tabular}{cc|c|ccc|ccc}
		\hline
		\multirow{2}{*}{+PSA} & \multirow{2}{*}{+FPC} &  \multirow{2}{*}{3D mAP (\%)} & \multicolumn{3}{c}{Ped. 3D AP (\%)} \vline &  \multicolumn{3}{c}{Cyc. 3D AP (\%)} \\
		& & & E. & M. & H. & E. & M. & H. \\
		\hline
		\hline
		&                     & 67.70 & 64.01 & 57.34 & 51.12 & 92.10 & 73.03 & 68.53\\
		\checkmark &            & 70.43 & \textbf{68.72} & 62.13 & 55.29 & 92.06 & 74.18 & \textbf{70.19}\\
		& \checkmark          & 70.19 & 66.98	& 61.06	& 55.93 & \textbf{93.64} & \textbf{74.55} & {69.97}\\
		\checkmark & \checkmark & \textbf{70.81}& {68.35} & \textbf{63.09} & \textbf{57.09} & {92.23} & {74.11} & {69.97}  \\
		\hline
	\end{tabular}
}
\end{table}

\subsubsection{{Fusion Ratio of FPC}}

According to formula \ref{eq:fusion}, the parameter $r$ determines the fusion ratio between the estimated centers and the segmented foreground points. To investigate the optimal fusion ratio of FPC, we conduct an experiment that compares the performance of PSA-Det3D using different $r$ to figure out the best fusion setting. The comparison results in Fig. \ref{fig:fusion_ratio} indicate that the fused confidences outperform the cases which use the estimated centers or foreground points only. It is reasonable to set the fusion ratio to $0.5$ for a balanced fusion between the foreground points and the centers. Although a large fusion ratio setting like $1.0$ can further improve the recall rate, the accuracy will be affected since too many candidate points are introduced.


\subsubsection{{Effect of Foreground Point Compensation}}

By comparing the first row and the third row of Table \ref{tab:ablation_ped_cyc}, we can observe that our FPC brings a remarkable improvement by 2.49\% on 3D mAP for small objects. It is noted that the FPC significantly boosts the 3D AP for ``pedestrian" by 2.97\%, 3.72\%, 4.81\% on three difficulties, where the improvement on ``moderate" and ``hard" pedestrians is greater than the ``easy" pedestrians. As the difficulty reflects the occlusion level, this result suggests that the proposed FPC can recognize the occluded objects and thus significantly improves the 3D AP on ``moderate" and ``hard" difficulties.



\section{Conclusion}
In this work, we propose a point-based network PSA-Det3D to alleviate the limitations of small object detection in 3D point cloud. The pillar set abstraction (PSA) and the foreground point compensation (FPC) are the core parts of PSA-Det3D. The PSA improves the query approach of SA with pillar query to learn informative features for small objects. The FPC module fuses the foreground points as well as the centroid information to recognize more occluded small objects during the proposal generation. The experiments on the KITTI 3D detection benchmark show that the proposed method outperforms existing point-based algorithms for small object detection.




\bibliographystyle{elsarticle-harv}
\bibliography{refs}



\end{document}